\documentclass[10pt,twocolumn,letterpaper]{article}

\usepackage{cvpr}              

\def\cvprPaperID{*****} 
\def\confName{CVPR}
\def\confYear{2022}

\usepackage{graphicx}
\usepackage{amsmath}
\usepackage{amssymb}
\usepackage{booktabs}

\usepackage[pagebackref,breaklinks,colorlinks]{hyperref}

\usepackage{microtype}
\usepackage{graphicx}
\usepackage[utf8]{inputenc} 
\usepackage[T1]{fontenc}    
\usepackage{hyperref}       
\usepackage{url}            
\usepackage{booktabs}       
\usepackage{amsfonts}       
\usepackage{nicefrac}       
\usepackage{microtype}      
\usepackage{xcolor}         
\usepackage{enumitem}

\usepackage{xspace}

\newcommand{\x}{\mathbf{x}}
\newcommand{\y}{\mathbf{y}}
\newcommand{\z}{\mathbf{z}}
\newcommand{\stb}{short-term behavior }
\newcommand{\Stb}{Short-term behavior }
\newcommand{\ltb}{long-term behavior }
\newcommand{\Ltb}{Long-term behavior }
\newcommand{\alg}{\texttt{BAMS}\xspace}
\newcommand{\algp}{\texttt{(BAMS)}\xspace}

\begin{document}

\title{Learning Behavior Representations Through Multi-Timescale Bootstrapping}

%
\author{
    Mehdi Azabou$^1$, Michael Mendelson$^1$,  Maks Sorokin$^1$,  Shantanu Thakoor$^2$,\\ 
     Nauman Ahad$^1$,  Carolina Urzay$^1$,   Eva L.~Dyer$^1$\\ {\em $^1$Georgia Institute of Technology, $^2$Deepmind}
}

\maketitle

\begin{abstract}
Natural behavior consists of dynamics that are both unpredictable, can switch suddenly, and unfold over many different timescales. While some success has been found in building representations of  behavior under constrained or simplified task-based conditions, many of these models cannot be applied to free and naturalistic settings due to the fact that they assume a single scale of temporal dynamics.  
In this work, we introduce Bootstrap Across Multiple Scales \algp, a multi-scale representation learning model for behavior: we combine a pooling module that aggregates features extracted over encoders with different temporal receptive fields, and design a set of latent objectives to bootstrap the representations in each respective space to encourage disentanglement across different timescales.
We first apply our method on a dataset of quadrupeds navigating in different terrain types, and show that our model captures the temporal complexity of behavior. We then apply our method to the MABe 2022 Multi-agent behavior challenge, where our model ranks 3rd overall and 1st on two subtasks, and show the importance of incorporating multi-timescales when analyzing behavior. 
\end{abstract}

\section{Introduction}

Until only recently, the study of animal behavior has been constrained to tasks and controlled experiments in which behavior is simple, universal and well understood \cite{tanigawa2022decoding, inagaki2022midbrain,dyer2017cryptography}.
Recent developments in animal tracking \cite{deeplabcut,lauer2022multi,SCHNEIDER2022} have made it possible to scale behavioral analysis to large video datasets that span complex and naturalistic behaviors. 
The next frontier is building tools that can quantify and describe such behavior. Some success was found in analyzing spontaneous behavior, but for specialized applications where restrictive assumptions are made: for example, spatially by restricting the analysis to a single limb of the animal \cite{shi2021learning}, or temporally by identify stereotyped modules of behavior only at the sub-second scale \cite{hsu2021b,wiltschko2015mapping}.

Behavior is not unimodal in naturalistic settings, it is rich and complex, it unfolds over many timescales in parallel \cite{colgan1978quantitative}: At the sub-minute scale, a mouse might be running, grooming or chasing another mouse, while at a longer timescale, it might cycle between different behavioral patterns depending on the time of day or mood. At an even longer timescale, the behavior can have a constant component characterized by inherent attributes like the morphology or age.

It is often hard to categorize the behavioral patterns or to identify subtle differences between them. A promising solution is to build models of the behavior in an unsupervised manner, by finding representations that capture the underlying state of an individual. Recent work in this direction utilizes contrastive learning objectives \cite{task_programming} or reconstruction-based objectives \cite{co2018self, CHEN2021332} to build such representations. However, these methods do not explicitly take into account the multi-timescale nature of behavior.

In this work, we introduce Bootstrap Across Multiple Scales \algp, a novel method for extracting representations that drive behavior over multiple timescales.
To do so, we design a temporal pyramid pooling module that aggregates information from multiple scales using different feature extractors with progressively larger receptive fields. Each feature extractor produces an embedding that encodes information from a particular scale. 
A novel contribution of our work is the bootstrapping of these embeddings in their respective spaces. In each timescale-specific embedding space, we use a latent predictive objective \cite{grill2020bootstrap} to bring the representations of neighboring time points closer to one another. What constitutes "neighboring time points" is going to depend on the timescale of the space we are working in: for short-term embeddings, we consider points that are within a few seconds of each other, while for long-term embeddings, we use a window in the order of minutes.
We also design a set of pretext tasks, related to both individual behavior and social interactions, that help guide the learning of meaningful behavior representations.

To test our approach, we first generate synthetic datasets from different robots with varying kinematics and body morphologies and demonstrate that our method can effectively build dynamical models of behavior that reveal both the type of agent the kinematics are generate from as well as the type of terrain the agent is walking on. After validating our approach on this synthetic dataset, we apply it to a multi-agent mouse behavior benchmark and challenge \cite{AIcrowd}, where we achieve competitive results that place us 3rd overall on the leaderboard and top-1 on two of the 13 subtasks. Our results demonstrate that our architecture and multi-timescale bootstrapping strategy can capture the temporal complexity of behavior and help build insight into the multi-timescale factors that modulate it.


\begin{figure*}[t!]
\centering
   \includegraphics[width=0.95\textwidth]{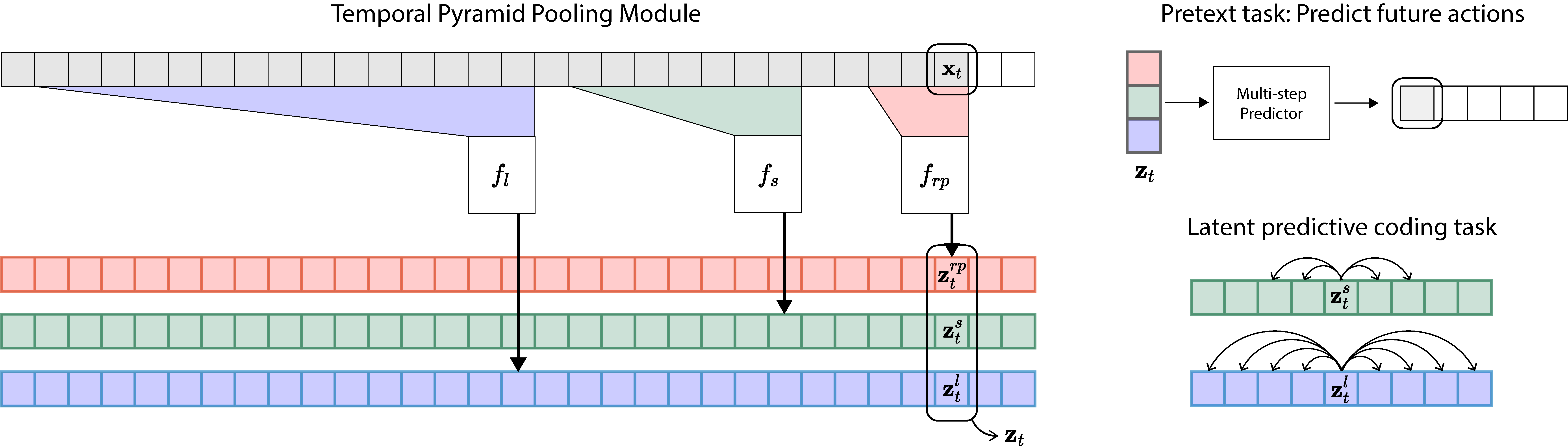}
   \caption{{\em Overview of our proposed \alg method}. The Temporal Pyramid Pooling Module, shown on the left, consists of 3 TCN encoders (recent-past encoder (red), \stb encoder (green), \ltb encoder (blue)), each with a different receptive field size. On the right, we illustrate the future action prediction, pretext task that the model uses to learn the representations as well as the multi-timescale latent prediction that uses different bootstrapping ranges depending on the timescale. \label{fig:overview}}
   \vspace{-3mm}
\end{figure*}

\section{Method}
In this section, we introduce our architecture and learning objectives, and show how we build multi-timescale representations that are separated explicitly through architectural choices and implicitly through diverse latent predictive losses.

\subsection{Model}

{\bf Temporal Convolutional Network.}
The building block of our model is a temporal convolutional network (TCN) \cite{bai2018tcn}  that takes in temporal sequences $\x_1,\cdots,\x_T$ as input for T time steps and outputs representations $\z_1,\cdots,\z_T$ for each time step (i.e. no temporal downsampling). 
TCNs use causal convolutions \cite{oord2016wavenet}, which means that the outputs at time $t$ depend only on inputs from time $t$ and earlier. The receptive field of the TCN encoder determines how much of the past the representation $\z_t$ encodes.
To expand the receptive field, or observable history, we stack more convolutions, making the network deeper. We can also opt for convolutions that are dilated at different rates \cite{chen2018encoder}.

{\bf Temporal Pyramid Pooling.}
In recurrent models with multiple memories, like LSTMs \cite{hochreiter1997long}, a gating mechanism is typically used to update the short and long-term memories, but the time horizons these memories encode are not explicit and are learned during training \cite{mahto2020multi}. In our model, we architecturally separate the different timescales by using TCNs with different receptive fields, that aggregate information at these different scales.
Inspired by spatial pyramid pooling architectures \cite{zhao2017pyramid,chen2018encoder}, 
our model uses three TCN branches, with different depths and dilation rates, forming a temporal pyramid pooling module:
\begin{itemize}
    \itemsep0em 
    \item {\bf Recent-past encoder,} $f_{rp}$: captures sub-second high-frequency information about the movement.
    \item {\bf \Stb encoder,} $f_{s}$: captures short-term dynamics (1sec-10sec) and targets momentary behaviors such as drinking, running or chasing. 
    \item {\bf \Ltb encoder,} $f_{l}$. captures long-term dynamics (minutes, hours) and targets longstanding factors that modulate behavior.
\end{itemize}

{\bf Predictor.}
All feature embeddings extracted by the TCNs are concatenated, to produce a multi-timescale behavior embedding,
\begin{equation}
    \z_t = \mathbf{concat}[\z^{rp}_t, \z^{s}_t, \z^{l}_t]
\end{equation}
The behavior embedding is fed to a multi-layer perceptron (MLP) predictor which solves different pretext tasks.

\subsection{Objectives}
We train our model by solving pretext tasks $\mathcal{L}_p$, and hypothesize that by doing so, good representations must be learned by the model. Furthermore, we encourage the model to separate features at different timescales by using a latent predictive loss $\mathcal{L}_r$. The final loss is:
\begin{equation}
\mathcal{L} = \mathcal{L}_p + \alpha \mathcal{L}_r
\end{equation}
where $\alpha$ is a hyperparameter.

{\bf Pretext task 1: Predicting future actions.}
Given observations $\x_t$ of the behavior at timesteps $0$ through $t$, the model extracts a feature embedding $\z_t$, and predictor $g$ is tasked to predict future actions $\y_{t+1}, \cdots, \y_{t+L}$, where $\y_t$ typically includes a subset of the observation features that $\x_t$ has.
\begin{equation}
    \mathcal{L}_{p, 1} = \|g(\z_t) - \mathbf{concat}[\y_{t+1}, \cdots, \y_{t+l}]\|_2^2
\end{equation}

By learning to predict future actions, the network learns to model the behavior dynamics.

{\bf Pretext task 2: Predicting hidden features.}
Another way of encouraging the learning of useful features is to have the model predict features that we hide from it. Given observations $\x_t$ of the behavior at timesteps $0$ through $t$, the model extracts feature embedding $\z_t$ and predictor $h$ is tasked to predict hidden features $\x_{h,t}$ from the same timestep.
\begin{equation}
    \mathcal{L}_{p, 2} = \|h(\z_t) - \x_{h,t}\|_2^2
\end{equation}

{\bf Multi-timescale latent predictive loss.}
As motivated above, our goal is to build representations that drive behavior at different timescales. We draw inspiration from recent work \cite{grill2020bootstrap, brave}, and use predictive regression losses, which are simpler and less computationally demanding compared to contrastive losses.
Given temporally neighboring \stb embeddings $\z^s_{t}$ and $\z^s_{t+\delta}$, we use a predictor $q_s$ that takes in embedding $\z^s_{t}$ and learns to regress $\z^s_{t+\delta}$ using the following loss:
\begin{equation}
    \mathcal{L}_{r,s} = \left \|\frac{q_s(\z^s_{t})}{\|q_s(\z^s_{t})\|_2} - \mathrm{sg}\left [ \frac{\z^s_{t+\delta}}{\|\z^s_{t+\delta}\|_2} \right] \right \|_2^2
\end{equation}
where $\mathrm{sg}[\cdot]$ denotes the stop gradient operator. Differently from \cite{grill2020bootstrap}, we do not use an exponential moving average of the network, but simply increase the learning rate of the predictor as done in \cite{brave}.

We use a similar setup for the \ltb embedding, where predictor $q_l$ is trained using the following loss:
\begin{equation}
    \mathcal{L}_{r,l} = \left \|\frac{q_l(\z^l_{t})}{\|q_l(\z^l_{t})\|_2} - \mathrm{sg}\left [ \frac{\z^l_{t+\delta}}{\|\z^l_{t+\delta}\|_2} \right] \right \|_2^2
\end{equation}
The critical difference between these two latent predictive terms is in the temporal augmentation being applied. For \stb embeddings, we randomly select samples, both future or past, that are within a small window $\Delta$ of the current timestep $t$, in other words, $\delta \in [-\Delta, \Delta]$. For \ltb embeddings which are to be stable at the level of a sequence, we sample any other time point in the same sequence, i.e. $t+\delta \in [0, T]$.

\begin{figure*}[t!]
\centering
   \includegraphics[width=0.85\textwidth]{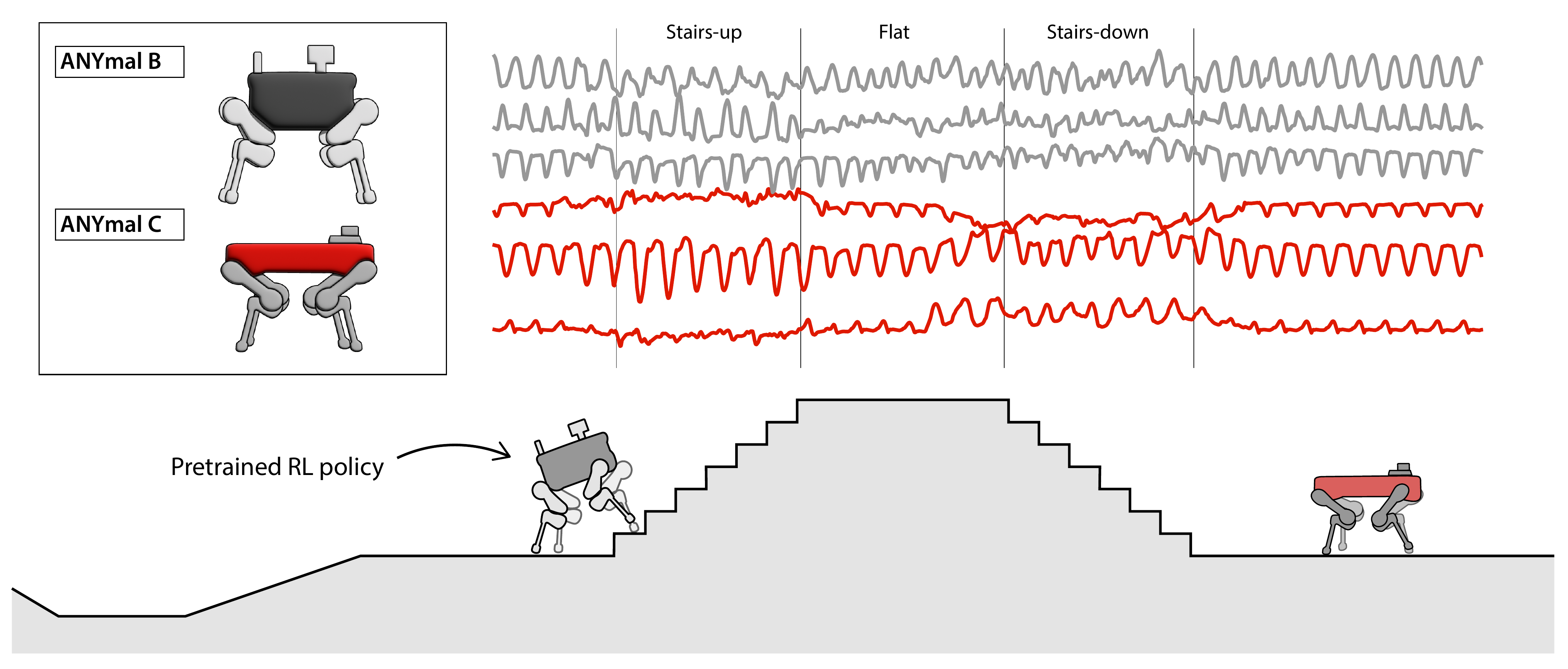}
   \caption{ 
   {\em 
   Quadrupeds walking on procedurally generated map.}
   (a) ANYmal B and ANYmal C have different morphology. We plot of subset of the tracked data as the robots traverse different terrains, and note a shift in kinematics for each terrain as well as a difference between robots.  
   (b) Illustration showing the robots walking on the procedurally generated map, with segments of different terrain types.
   \label{fig:robots}}
   \vspace{-3mm}
\end{figure*}

\section{Results}

\subsection{Simulated Legged Robots Experiment}

To demonstrate our model's ability to separate behavioral factors that play out over different timescales, we collect data from multiple simulated robots walking on different terrains. We use NVIDIA’s Isaac Gym \cite{isaacgym} simulation environment, and train reinforcement learning agents to walk on challenging terrains using \cite{rudin2022learning}, which enables massive parallelism on small workstations. Specifically, we train two different robots, ANYmal B and ANYmal C, differentiated by their morphology. We generate 76 sequences of these robots walking through a procedurally generated map composed of multiple segments of different terrains types (Figure~\ref{fig:robots}). There are 5 terrain types we consider: stairs (up), stairs (down), slope (up), slope (down), and flat. 
The way the robot walks will depend on its morphology as well as the type of terrain it is walking on. We designed this data to simulate the multi-timescale nature of behavior: morphology as a factor effects behavior at a global scale and terrain type does so at a shorter timescale.

We record these robots for approximately one minute and target 12 features, including the: linear and angular velocity of the robot's body, as well as the position and velocity readings from the thigh, hip, and shank from each of the 4 robot legs.
We split the dataset into train/test sets (80/20), and report linear decoding accuracies reflecting the decodability of terrain type and robot type from the learned behavior embeddings. 

Results in Table~\ref{tab:robots} compare the representational quality of each of the behavior representations both combined and individually. The results suggest that terrain type manifests over both timescales, while robot type is better decoded from the long-term behavior representation alone.

\begin{table}[t!]
\centering
\caption{{\em  Results on robot behavioral data.}  For each task (terrain and morphology), we report the linear decoding accuracy for our model when we use all representations in the model (full space), only the short-term embedding, and only the long-term embedding.}
\begin{tabular}{lcc}
\hline
      & \multicolumn{2}{c}{F1-score} \\
Model & Terrain type & Robot type \\ \hline
Short-term + Long-term  & {\bf 0.73} & 0.98 \\ 
Short-term only & 0.50 & 0.86 \\
Long-term only  & 0.62 & {\bf 0.99 } \\ \hline
\end{tabular}
\label{tab:robots}
\end{table}

\subsection{Experiments on the MABe Multi-agent Behavior Dataset}

\subsubsection{Experimental Setup and Tasks}

\paragraph{Dataset description.}
The mouse triplets dataset is part of the Multi Agent Behavior Challenge (MABe 2022) \cite{AIcrowd}. The goal of the challenge is to build representations of the data that capture the social behavior of the mice.

The dataset consists of 1600 one-minute recordings of trios of interacting mice. Over the course of these sequences, the mice might exhibit individual and social behaviors. Some might unfold at the frame level, like chasing or being chased, others at the sequence level, like light cycles affecting the behavior of the mice or mouse strains that inherently differentiate mice behavior. 13 (private) sets of labels describing all these factors are used to evaluate the quality of learned representations. These include both frame-level and sequence-level labels.

The poses of the mouse triplet are recorded at 30 fps, and each mouse's pose is described by 12 keypoints extracted from top-view videos.
The public training set includes these keypoints, which account for a total of more than 2.8M pose estimates.

\begin{table*}[t!]
\centering
\caption{
     \label{tab:mouseacc} 
     {\em  Linear readouts of mouse behavior.} We report the top 4 models on the leaderboard of the unsupervised behavior embedding challenge. The scores show the performance of the linear readouts from the representation across 13 different tasks. This evaluation metric is MSE in the case of tasks 1 and 2 (indicated by *), since they are continuously labeled. In the rest of the subtasks, which are binary (yes/no), F1-scores are used. The best-performing models are those with low MSE scores and high F1-scores.
     }
\vspace{0.1in}
\resizebox{0.99\textwidth}{!}{
\vspace{-2mm}
\begin{tabular}{l|cc|cccc|ccccccccc}
& & & \multicolumn{4}{c|}{Sequence-level subtasks} & \multicolumn{9}{c}{Frame-level subtasks} \\
Model & F1-score & MSE & T1$^*$ & T2$^*$ & T3 & T13 & T4 & T5 & T6  & T7  & T8 & T9  & T10  & T11 & T12 \\   
\hline
\# 1 & {\bf 30.3} & {\bf 0.09296}   & 0.09019 & {\bf 0.09523} & {\bf 82.20} & 69.40 & 1.90  & 1.24 &  {\bf 71.62} & 55.52 & {\bf 30.20} & 0.40 & 1.63 & 1.10 & {\bf 20.45} \\ 
\# 2 & 28.3 & 0.09289   & 0.09057 & {\bf 0.09513} & 67.20 & 66.90 & {\bf 2.70}  & {\bf 6.60} &  71.47 & {\bf 54.67} & 20.30 & {\bf 0.68} & {\bf 3.31} & {\bf 2.37} & 18.54\\ 
\# 3 \alg (Ours) & 28.4 & 0.09298 &  {\bf 0.09037} & 0.09513 &  67.10  & {\bf 69.50}  &  2.16 & 2.31 & 66.42  & 53.28  & 30.18  & 0.45 & 1.65 & 1.14 & 19.14 \\ 
PCA baseline & 7.99 & 0.09430 &  0.09415 & 0.09449 &  33.83  & 4.13  &  0.00 & 0.00 & 12.69  & 0.08  & 0.00  & 0.00 & 0.00 & 0.00 & 0.00  \\ 
 \hline
\end{tabular}
}
\end{table*}

\begin{figure*}[t!]
\centering
   \includegraphics[width=0.85\textwidth]{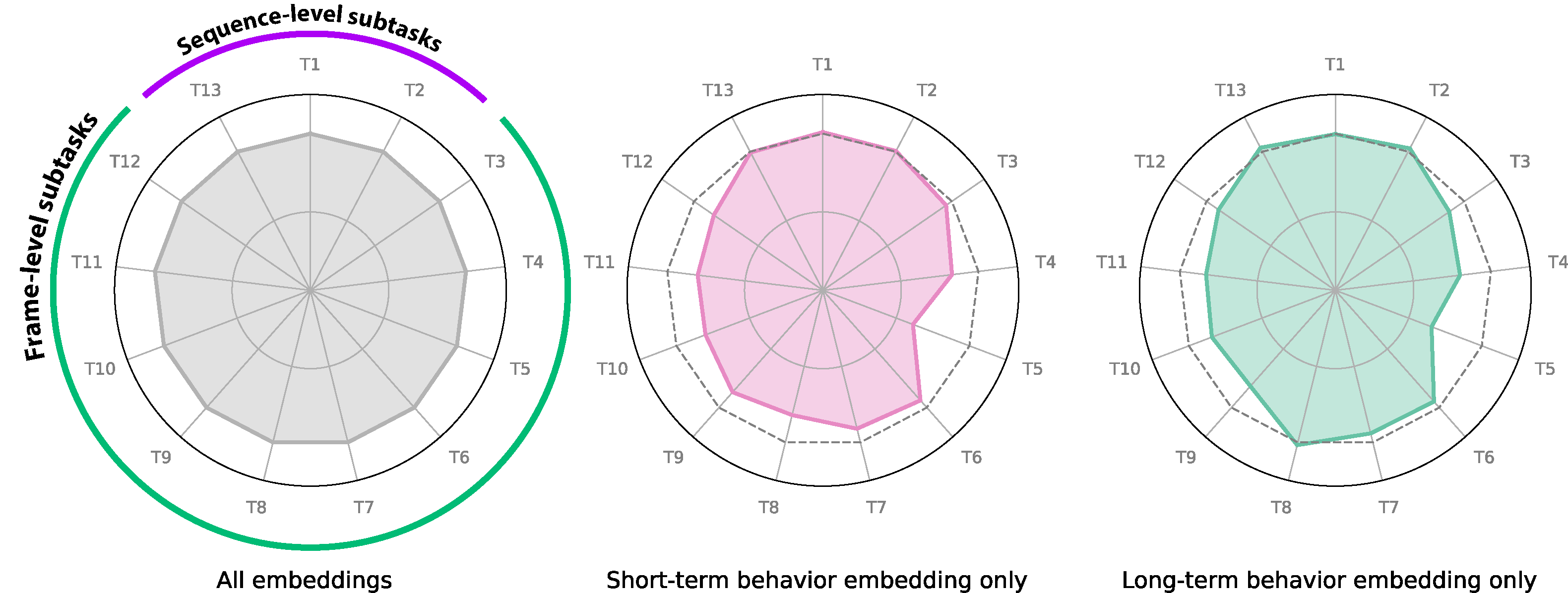}
   \caption{{\em Relative change (in \%) of subtask scores from the full multi-timescale embedding (gray)}. We show the change in performance when reading out labels from the short-term behavior embedding only (pink) or the long-term behavior embedding only (green).}  \label{fig:ablations}
   \vspace{-3mm}
\end{figure*}

\vspace{-2mm}
\paragraph{Training protocol.}
We process the tracking data to extract 36 features characterizing each mouse individually, including head orientation, body velocity and joint angles. Each mouse is processed independently by the TCN feature extractor to generate, at time t, embeddings $\z_{t,1}$, $\z_{t,2}$ and $\z_{t,3}$ respectively. The Temporal Pyramid Pooling module, composed of the recent-past encoder, the \stb encoder and the \ltb encoder have output sizes of 16, 32 and 16 and receptive fields of 3, 30 and 253 frames respectively.

As a pretext task, we select 6 of the input features to be prediction targets for future timesteps. We train the network to predict $L=15$ timesteps into the future, which amounts to $0.5$ seconds. At this stage, we do not incorporate any information about how the mice are interacting when solving this future action prediction task.

To model the animal interactions, we train the model to also predict the distances between the trio at time $t$. These distances are hidden, and the input features do not include any information about the global position of the mice in the cage, so the model can only rely on the inherent behavior and movement of each individual mouse to draw conclusions about their level of closeness. We build a network $h$ that takes in the embeddings of two mice $i$ and $j$ and predicts the  distance $d_{i,j}$ between them. 
\begin{equation}
    \mathcal{L}_{p, 2} = \|h(\z_{t,i}, \z_{t,j}) - \mathrm{d}_{i,j}\|_2^2
\end{equation}

Finally, we include the latent predictive coding objective with parameters $\Delta=5$ and $\alpha=0.1$. The model is trained for 2000 epochs using the Adam optimizer with a learning rate of $10^{-3}$.

\vspace{-2mm}
\paragraph{Evaluation protocol.}
To evaluate the representational quality of our model, we take the provided test set and compute representations $\z_{t,1}$, $\z_{t,2}$ and $\z_{t,3}$ for each mouse respectively. We aggregate these three embeddings into a single mouse triplet embedding using two different pooling strategies. First, we apply average pooling to get $\x_{t, \mathrm{avg}}$. Second, we apply max pooling and min pooling, then compute the difference to get $\x_{t, \mathrm{minmax}}$. Both aggregated embeddings are concatenated into a 128d embedding which we use for our submission.

We submit the embeddings to the AIcrowd platform, where evaluation is performed and averaged over 3 random data splits. For each one of the 13 tasks, a linear layer is trained on top of the frozen representation, producing an F1 score or a mean squared error. The averages of these metrics produce a final score F1-score and MSE error that indicates the overall performance of the model.

\subsubsection{Results}
Our model achieves competitive results on the MABe 2022 challenge \cite{AIcrowd}, as can be seen in Table~\ref{tab:mouseacc}. We rank 3rd globally, and 1st on two of the 13 subtasks, while remaining competitive on most other subtasks. 

To quantify the individual role played by each of the behavioral embeddings at different timescales, we perform linear evaluation on top of each embedding separately, as shown in Figure~\ref{fig:ablations}. First, we note that the performance on subtasks T1, T2 and T13 is maintained, suggesting that both short-term and long-term behavior embeddings might contain information about the sequence-level subtasks. Second, we find subtasks where performance decreases for one embedding but is maintained for the other, like tasks T3 and T8, this suggests that the corresponding behavior might mainly manifest in one timescale. Finally, we note a number of tasks, like T4 and T5 where performance decreases for both embeddings, suggesting that the corresponding behavior manifests across both short and long timescales. Further investigation will be performed, once a description of the tasks is released.

\section{Related Work}

\subsection{Animal behavior analysis}

Most pipelines for animal behavior consist of three key steps \cite{luxem2022video}: 
(1) pose estimation \cite{deeplabcut,sun2021self,wu2020deep}, (2) spatial-temporal feature extraction \cite{luxem2020identifying, CHEN2021332}, and (3) quantification and phenotyping of behavior \cite{Nair2022.04.19.488776, MARSHALL2022102522, wiltschko2020revealing}. 
In our work, we consider the analysis of behavior after pose estimation or keypoint extraction is performed. However, one could imagine using the same ideas for representation learning being used in video analysis, where self-supervised losses have been proposed recently for keypoint discovery \cite{sun2021self}.

\vspace{-2mm} 
\paragraph{Disentanglement of animal behavior in videos.}~In recent work \cite{shi2021learning}, disentangled behavior embeddings (DBE) are learned from video by separating non-behavioral features (context, which recording condition etc) from the dynamic behavioral factors (pose). 
This is performed by learning two encoders and giving either multi-view dynamic information or a single image to the two encoders. 
In comparison to this work, our paper considers the separation of behavior across different timescales and considers the construction of a global embedding that will be consistent over long timescales.

\vspace{-2mm} 
\paragraph{Modeling social behavior.}~ For social and many-animal datasets, there are a number of other challenges that arise. Simba \cite{nilsson2020simple} and MARS  \cite{segalin2021mouse} have similar overall workflows for detecting keypoints and pose of many animals and classifying social behaviors. More recently, a semi-supervised approach TREBA has been introduced \cite{sun2021task} for building behavior embeddings using task programming. TREBA is built on top of the trajectory VAE \cite{co2018self}, a variational generative model for learning tracking data. In our work, we do not consider a reconstruction objective but a future prediction objective, in addition to bootstrapping the behavior representations at different timescales.

\subsection{Representation learning for sequential data}

The self-supervised learning (SSL) framework has gained a lot of popularity recently due to its impressive performance in many domains with a wide range of applications. BYOL \cite{grill2020bootstrap} proposes a framework in which augmentations of a sample are brought closer together in the representation space through a predictive regression loss. Recent work \cite{myow,brave,guo2020bootstrap,niizumi2021byol,azabouusing} applys BYOL to learn representations of sequential data. In such cases, neighboring samples in time are considered to be positive examples of each other, assuming temporal smoothness of the semantics underlying the sequence. The model is trained such that neighboring samples in time are mapped close to each other in the representation space. 

The idea of using multi-scale feature extractors can be found in representation learning. In \cite{brave}, a video representation learning framework, two different encoders process a narrow view and a broad view respectively. The narrow view corresponds to a video clip of a few seconds, while the broad view spans a larger timescale. The objective, however, is different to ours, as the narrow and broad representations are brought closer to each other, in the goal of encoding their mutual information. This strategy is also used in graph representation learning where a local-neighborhood of node is compared to its global neighborhood \cite{hassani2020contrastive, hu2020graph}. Our method differs in that we bootstrap the embeddings at different timescales separately, this is important to maintain the fine granularity specific to each timescale, thus revealing richer information about behavioral dynamics.

\section{Conclusion}
Variability of animal behavior is likely to be driven by a number of factors that can unfold over different timescales. Thus, having ways to model behavior and discover differences in behavioral repertoires or actions across timescales could provide insights into individual differences, and help, for example, detect signatures of cognitive impairment \cite{wiltschko2020revealing}.
We make steps towards  addressing these needs by proposing a novel approach that learns representations for behavioral data at different timescales. 

Experiments on synthetic robot datasets and real-world mice datasets show that our method can learn to encode the wide temporal spectrum of behavior representations. This makes it possible to distinguish global as well as temporally local behaviors. This separation allows us to better discover  variations at different timescales and across different agents, helping us to develop better insights. In the future, we plan to test our model on additional datasets to further study how behavior patterns are exhibited at different frequencies.

\section*{Acknowledgements}
We would like to thank Mohammad Gheshlaghi Azar and  Remi Munos for their feedback on the work. This project was supported by NIH award 1R01EB029852-01, NSF award  IIS-2039741, as well as generous gifts from the Alfred Sloan Foundation, the McKnight Foundation, and the CIFAR Azrieli Global Scholars Program.

\nocite{*} 

\bibliography{main}
\bibliographystyle{ieeetr}

\end{document}